# Macroscopic Emission Modeling of Urban Traffic Using Probe Vehicle Data: A Machine Learning Approach

Mohammed Ali El Adlouni, Ling Jin, Xiaodan Xu, C. Anna Spurlock
*Lawrence Berkeley National Laboratory*
Berkeley, CA

Alina Lazar
*Youngstown State University*
Youngstown, OH
*Berkeley National Laboratory*
Berkeley, CA

Kaveh Farokhi Sadabadi
University of Maryland

Mahyar Amirgholy,
*Kennesaw State University*
Marietta, GA

Mona Asudegi
*Federal Highway Administration*
Washington DC, D.C. 20590
mona.asudegi@dot.gov

*Abstract*—Urban congestions cause inefficient movement of vehicles and exacerbate greenhouse gas emissions and urban air pollution. Macroscopic emission fundamental diagram (eMFD) captures an orderly relationship among emission and aggregated traffic variables at the network level, allowing for real-time monitoring of region-wide emissions and optimal allocation of travel demand to existing networks, reducing urban congestion and associated emissions. However, empirically derived eMFD models are sparse due to historical data limitation. Leveraging a large-scale and granular traffic and emission data derived from probe vehicles, this study is the first to apply machine learning methods to predict the network-wide emission rate to traffic relationship in U.S. urban areas at a large scale. The analysis framework and insights developed in this work generate data-driven eMFDs and a deeper understanding of their location dependence on network, infrastructure, land use, and vehicle characteristics, enabling transportation authorities to measure carbon emissions from urban transport of given travel demand and optimize location-specific traffic management and planning decisions to mitigate network-wide emissions.

## I. Introduction

The transportation sector is a major contributor to U.S. energy use and greenhouse gas emissions. When travel demand exceeds the infrastructure capacity, road network becomes congested with vehicles spending more time on the road in a stop-and-go motion and making frequent acceleration and deceleration, intensifying emissions per vehicle miles traveled.

The macroscopic fundamental diagram (MFD) describes an orderly and consistent relationship between average vehicle flow and average traffic density when both are measured across a certain urban network [1][2][3]. The theoretical underpinnings of the MFD suggest that consistent network-level relationships may also exist for driving cycle parameters that are most closely associated with emissions. Several studies have demonstrated, using microsimulation or empirical data, that it was feasible to estimate emissions based on aggregated traffic characteristics [4][5][6]. The emission macroscopic fundamental diagram (eMFD), that captures a consistent relationship among emission and aggregated traffic variables, has since emerged as an efficient tool for real-time monitoring of region-wide emissions and optimal allocation of travel demand to existing networks, reducing urban congestion and associated emissions.

However, empirically derived eMFD models are sparse in the literature because quantifying such macroscopic relationship needs data aggregated from link level traffic and emission data that are historically limited. Probe vehicle data provides high spatiotemporal coverage of traffic across different networks by collecting traffic variables from vehicle's precise location, speed, and direction. When combined with emission models, the traffic activity data derived from the probe vehicles can be used for modeling emissions to traffic relationship and determine its dependence on locations at scale.

## II. Data and Method

This study collected three months of HERE probe data (Sept - Nov) in 2019 for the full U.S. including network geometry, traffic counts, speeds, and number of probes. Traffic volume, density, and speed at 15-minute granularity are first estimated at road segments, then aggregated to the network level for individual U.S. urban census tracts [7]. Tract-level vehicle running exhaust emission rates are aggregated from link level emissions estimated by coupling the link-level vehicle activities with MOVES-Matrix emission rates [8] for four states: New York, Colorado, Texas, and Georgia. MOVES-Matrix is a pre-generated MOVES emission rate database, that can be queried for given species as a simple function of vehicle type, vehicle vintage, road type, and link average speed.

Vehicle characteristics and location factors representing various transportation supply and demand characteristics (following [7]) are included to predict tract-level emissions at a given traffic density. Four machine learning methods (Random Forest, XGBoost, Support Vector Machine, and LightGBM) are evaluated. We apply TreeExplainer to interpret the interactions of the location and vehicle characteristics with traffic density on influencing the emission to density relationship (i.e. eMFD). An overview of the data and analysis process is illustrated in Fig. 1.

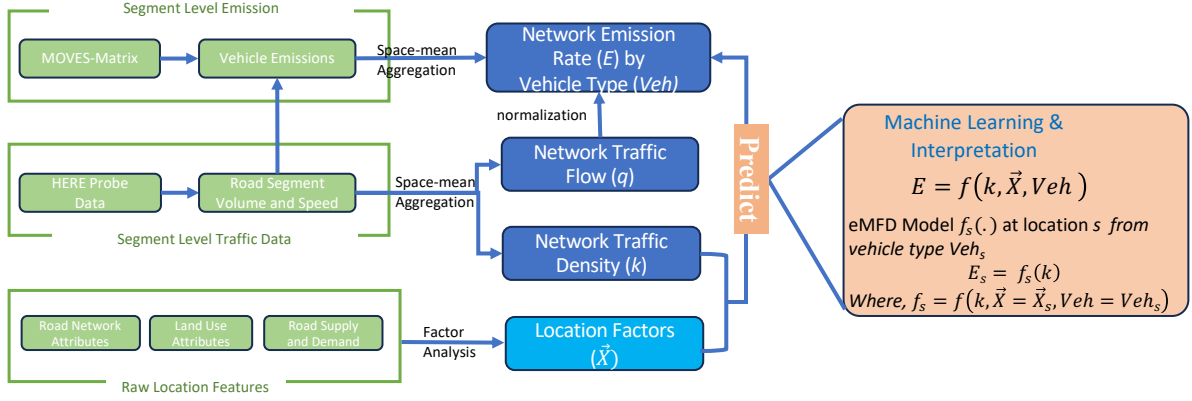

Fig. 1. Overview of data process and analysis of the paper.

## III. RESULTS AND DISCUSSIONS

### A. Aggregated Network Traffic and Emission Rates

The road segment level HERE and emission data are aggregated to derive flow, density, and per vehicle mile $CO_2$ emission rates in 13,022 individual census tracts from the four selected states at a 15 minute interval. The tract-level median density and emission rates at 5:00PM is illustrated for New York City in Fig. 2. Higher traffic density (Fig. 2a) appears in central business areas known for chronic congestion which also induces higher per mile emissions (Fig. 2b).

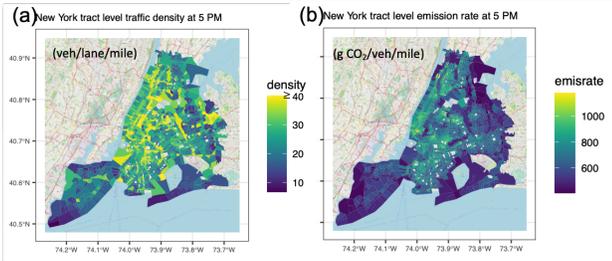

Fig. 2. (a) Tract level traffic density and (b) $CO_2$ emission rate aggregated from vehicle probe data in New York City.

At the network level, the aggregated flow and density demonstrate an orderly relationship (i.e. MFD) which varies by the location of the network. Tract 2 has lower network capacity (the maximum flow) than Tract 1 (Fig.3a), therefore has lower speed and is more easily to get congested, leading to higher emission rates of given traffic density (Fig. 3b). At the same time, emission rates vary by vehicle characteristics. Older fleet, represented by light duty vehicles of vintage year 2000 and older, is equipped with less efficient engines and therefore tends to emit higher $CO_2$ per mile traveled than newer fleet represented by vehicles of vintage year 2018 (Fig.3b).

### B. Performance of Machine Learning Models

We use location factors and two bounding light duty vehicle fleet types in urban areas (vehicles of vintage year 2000 and earlier v.s. vehicles of vintage year 2018), to predict network level emission rates at given traffic density. We split

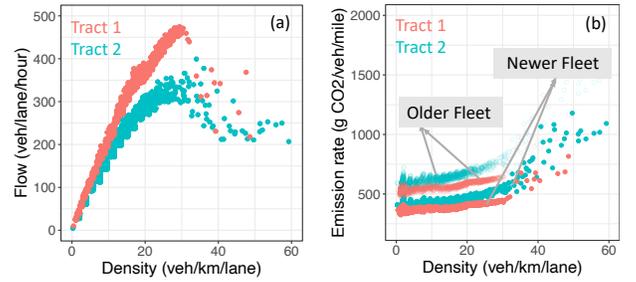

Fig. 3. Example (a) flow to density relationship (MFD) by location, and (b) emission rate to density relationship by location and fleet characteristics.

6,629,818 rows of data from the urban tracts into 80% training and 20% testing datasets. The model performance is evaluated using four metrics: $R^2$, mean absolute error (MAE), root mean squared error (RMSE), and mean absolute percentage error (MAPE) in TABLE I. Across all four metrics, XGBoost consistently shows the best performance among the machine learning models evaluated here.

TABLE I. MODEL PERFORMANCE METRICS ON 20%TESTING DATA

| ML Models | Performance Measures | | | |
|---|---|---|---|---|
| | $R^2$ | MAE (g/veh/mile) | RMSE (g/veh/mile) | MAPE (%) |
| XGBoost | 0.92 | 16.25 | 49.83 | 2.87 |
| Random Forest | 0.91 | 119.32 | 240.5 | 7.58 |
| LightGBM | 0.91 | 104.48 | 245.00 | 6.83 |
| Linear SVM | 0.82 | 171.46 | 345.03 | 15.42 |

### C. Interpretation

Using the interaction SHAP values coupled with the best performing XGBoost model, TreeExplainer uncovers the influence of location factors and vehicle characteristics on the emission to density relationships learned by the model. Fig. 4 presents the importance ranking of the input features according to their interaction SHAP values with density.

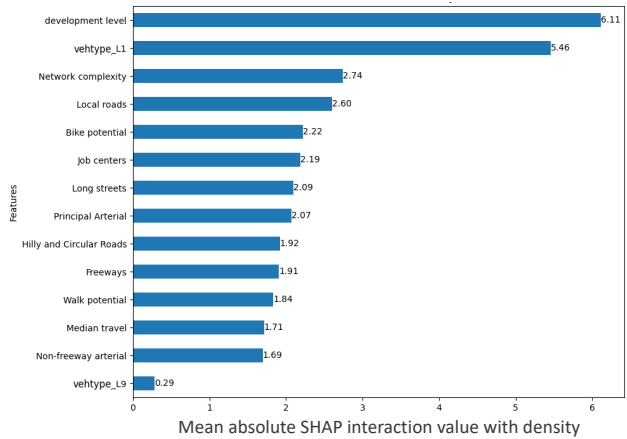

Fig. 4. Importance ranking of location and vehicle characteristics on eMFD based on SHAP interaction value with density in predicting emission rates.

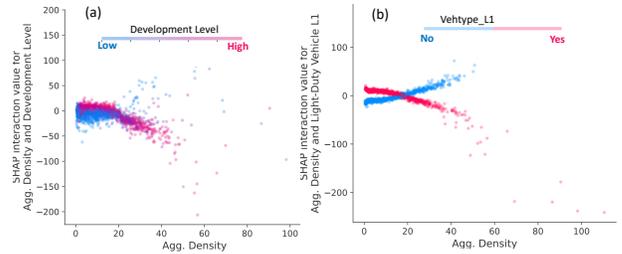

Fig. 5. SHAP interaction plots showing the emission dependence on density altered by the location factor "development level" (left), and vehicle type L1density (right).

The two top-ranking features are "development level" and "vehtype_L1". The "development level" factor captures smaller tracts characterized by higher intersection and street density, higher population and job density, and higher proportion of non-freeway arterials. The "vehtype_L1" feature represents the emission variation between the newer and older engines. Past studies typically focused on a single region and therefore derived eMFDs (i.e. emission to traffic relationship) accounting for vehicle types. This analysis has shown that location to location differences may vary the eMFD shapes in a comparable magnitude as that introduced by the fleet composition, underlying the importance in urban planning in emission controls in addition to clean vehicle technologies.

We can see "development level" and "vehtype_L1" both exhibit diverging effects between high and low factor scores *after* density value reaches 20 veh/lane/mile. Networks with higher development level tend to have lower emission rates of given density, likely due to their higher street density to maintain the flow in the congested regime [7]. The newer light duty vehicles operate more efficiently during high density conditions and therefore has lower emission rates relative to the older vehicle type (Fig. 5b).

## IV. Conclusions

This paper addressed research gaps on empirically derived and location-flexible macroscopic traffic emission models by leveraging a large-scale granular probe vehicle data combined with the state-of-art emission modeling in U.S. urban areas. Coupled with the best performing XGBoost model, the TreeExplainer uncovers influences of both location factors and vehicle fleet characteristics on the emission to traffic relationship. We found emission intensity of given traffic demand in an urban area is strongly dependent on its network typology, infrastructure, and land use attributes that underly their propensity for congestion. This finding underscores the importance of urban planning in urban traffic emission controls in addition to promoting clean vehicle technology.

The location dependent macroscopic relationships between emissions and traffic, following the analysis framework proposed in this study, can enable transportation authorities to measure carbon emissions from urban transport of given travel demand and optimize location-specific traffic management and planning decisions to mitigate network-wide emissions.

## Acknowledgment

This work is funded by the Federal Highway Administration (FHWA) Office of Transportation Policy Studies, under Interagency Agreement 693JJ318N300068 with the U.S. Department of Energy under Contract No. DE-AC02-05CH11231 to Lawrence Berkeley National Laboratory. Any statement or results in this paper reflects the authors' perspectives and not necessarily the point of view of FHWA.